# Design of an Ultrasound-Guided Robotic Brachytherapy Needle-Insertion System

Nikolai Hungr, Jocelyne Troccaz, *Member, IEEE*, Nabil Zemiti, and Nathanaël Tripodi

*Abstract*— In this paper we describe a new robotic brachytherapy needle-insertion system that is designed to replace the template used in the manual technique. After a brief review of existing robotic systems, we describe the requirements that we based our design upon. A detailed description of the proposed system follows. Our design is capable of positioning and inclining a needle within the same workspace as the manual template. To help improve accuracy, the needle can be rotated about its axis during insertion into the prostate. The system can be mounted on existing steppers and also easily accommodates existing seed dispensers, such as the Mick Applicator.

## I. INTRODUCTION

PROSTATE cancer is the most common cancer in France with 62 245 new cases estimated in 2005, and the second most common cancer in the US with 186 320 new cases estimated in 2008 [1] [2]. A reduction of death rates for prostate cancer has been reported for a number of western developed countries over the past 10 years. This has been attributed in part to improved diagnosis and treatment techniques [1].

Brachytherapy is a technique that has only recently become an important treatment method for specific cases of prostate cancer. It has been shown that brachytherapy is a reliable technique with a high success rate [3]. The technique involves the localized irradiation of the prostate gland by the insertion of about 100 radioactive seeds, each the size of a grain of rice. The seeds are placed in the prostate by means of hollow needles inserted through the perineum of the patient in the lithotomy position and using trans-rectal ultrasound (TRUS) guidance. A template, as seen in Fig. 1, is used to insert the needles along a grid of horizontal holes.

An important element for the success of a brachytherapy intervention is the uniform distribution of radioactive dose throughout the entire volume of the prostate, without overdosage and without affecting adjoining organs such as the bladder, rectum, seminal vesicles or urethra. The procedure is therefore heavily reliant on the ability of the clinicians and physicists in reproducing the pre-planned dosimetry within the prostate. Multiple limitations to the current manual technique make this a difficult task.

The primary difficulty lies in the mobility of the prostate and surrounding soft tissues during the intervention. Both the insertion of the needles and the movement of the TRUS probe cause significant motion and deformation of the prostate [4]. Since the dosimetry plan is typically based on the manual segmentation of at most two sets of ultrasound images taken before the insertion of the needles, the resultant accuracy of the seed placement is difficult to verify. This accuracy is additionally affected by a number of other factors, including the random migration of the seeds upon their release within the prostate, the flexion of the needles upon insertion and prostatic edema during the intervention.

Another important limitation to the technique is that needle insertion is restricted to the horizontal axes defined by the needle template. Not only is needle placement limited to a grid of 5mm spacing, but perhaps more importantly, this parallel grid system does not allow access behind the pubic arch in the relatively frequent case of the latter eclipsing parts of the prostate.

These issues, amongst others, result in a lengthy and unavoidably repetitive procedure that relies heavily on the experience of the clinicians and physicists. Many of the issues described above could, in fact, be solved with the use of a robotic system for the insertion of the needles. In this paper we present a prototype of such a system that is currently being developed for use at the Grenoble University

Manuscript received April 7, 2009. This work was supported in part by an ANR grant (Tecsan Program – Prosper project).

Nikolai Hungr and Jocelyne Troccaz are with the TIMC Laboratory, 38706 La Tronche cedex, France (phone: +33 (0)4 56 52 00 52; fax +33 (0) 4 56 52 00 55; nikolai.hungr@imag.fr; jocelyne.troccaz@imag.fr).

Nabil Zemiti was with the TIMC Laboratory, 38706 La Tronche cedex, France. He is now with the LIRMM Laboratory, UMR CNRS/UM2 5506, 34392 Montpellier, France (Nabil.Zemiti@lirmm.fr).

Nathanaël Tripodi was with the TIMC Laboratory, 38706 La Tronche cedex, France.

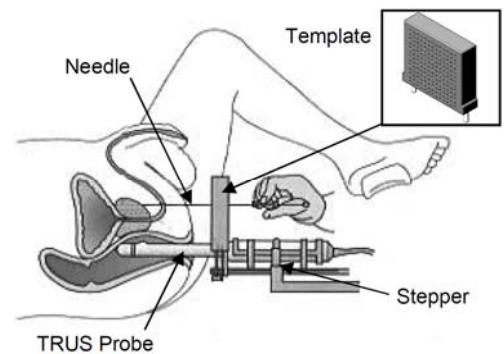

Fig. 1. Typical brachytherapy setup, showing the various components used in the procedure. From http://www.uropage.com/index.htm.

Hospital (CHUG).

## II. PRIOR ART

A number of ultrasound-guided transperineal brachytherapy robots have been developed to date. Each introduces a specific set of technological advances aimed at reducing the uncertainties found in the manual technique.

Perhaps the most complete system, in terms of automation of the various complex tasks involved in brachytherapy, is the system described in [5]. Their system includes a 9-DOF gross positioning system on which is mounted a 2-DOF TRUS probe driver and a 3-DOF x-y translation and pitch inclination gantry, which itself holds a 2-DOF needle and seed driver. The needle driver is also capable of rotating the needle cannula about its axis during insertion. A specific element studied by this group is the characterization of needle insertion and rotation velocity on the needle's interaction with soft tissue, showing that needle rotation can decrease tissue puncture force and deformation as well as needle deflection [6].

Reference [7] describes a robot that uses a gross positioning driver to initially position a needle before insertion and a smaller plunger system for the final insertion into the prostate. The system allows for the parallel insertion of needles with high accuracy and faster than with the manual technique. As in [5], the robot also allows for the rotation of the needle to minimize needle bending and insertion force, however it does not allow for needle inclination for pubic arch avoidance. The authors report that tissue movement during insertion is an important issue, with movements on phantoms being on the order of 5mm.

Another system, developed by [8], consists of a four-degree-of-freedom (DOF) robot that is capable of both parallel and inclined needle insertion, allowing for finer seed placement control as well as access behind the pubic arch. The system was designed as a direct replacement for the template, so needle insertion is manual. The design is therefore unable to counter the tissue movement and needle flexion issues encountered in the standard manual technique.

To date, the only robotic brachytherapy robot to have been tested clinically is the system described by [9]. The system uses two parallel planar motion stages to position and angle the needle through a pair of ball joints. Although its vertical position above the operating site restricts the clinician's field of view, the architecture is relatively compact and, if required, is easily replaceable with the manual template. As in the system developed by [8], it requires manual insertion of the needles and thus has no control on the needle-tissue interactions during insertion.

Numerous other designs exist, including the multiple-needle insertion robot in [10], the cable-driven parallelogram robot presented in [11], the industrial robot-guided template of [12] and several MRI- and CT-based designs of varying architectures [13]-[17]. The latter are, in general, non-standard techniques that require modified surgical procedures.

## III. DESIGN REQUIREMENTS

In designing our robotic brachytherapy needle-insertion system, we came up with a set of design constraints based on the various results from the literature as described above, upon a survey distributed to our team of clinicians as well as on our observation of brachytherapy interventions.

One of the main goals of the system was to provide benefits in the ease of the procedure and more importantly in its clinical success. For an experienced surgeon, the manual technique is relatively uncomplicated and generally provides excellent results for standard cases. The technique becomes difficult, however, in more complex cases such as patients with larger prostates (prostatic volume >50 $cm^3$ [18]), softer prostates that are more prone to deformation and seed migration, or patients with distinct anatomy such as increased musculature or a tight pubic arch. With this in mind, we based our design on the following design requirements:

- Rotation of needle about its axis during insertion: to minimize tissue and needle deformation.
- Needle pitch and yaw inclinations: to reach behind the pubic arch.
- Precision: <1 mm, including tissue-needle interaction effects.
- Ease of operation: the robot and the accompanying surgical procedure must be faster and no more complex than the manual technique.
- Minimal obstruction: the clinician's access and view of the perineum must not be restricted.
- Workspace: should be able to cover the 60 x 60 mm grid of the template, along with pitch and yaw inclinations.
- Weight: <5 kg, to ensure ease of installation and handling.
- Compatibility: compatible with existing steppers, needles and seed insertion tools (i.e. replace only the template).
- Safety: must ensure the safety of the patient and operating room staff. Must also be able to revert rapidly and easily back to the manual technique in case of an emergency.
- Sterilization: must meet regulations on operating room sterilization.

## IV. ARCHITECTURE

A CAD model of our prototype is shown in Fig. 2. The design consists of two primary elements: the needle positioning module and the needle insertion module. The two are independent of each other, allowing each to be modified or replaced separately if necessary. The needle positioning module can be mounted either to an existing stepper or to a custom stepper via an interchangeable set of mounting brackets. It is mounted on the lateral side of the stepper and thus liberates the entire space directly above and to the opposite lateral side of the operating site. The positioning module positions the needle along the appropriate insertion axis and the insertion module then drives the needle to a given depth. The clinician can then insert the seed.

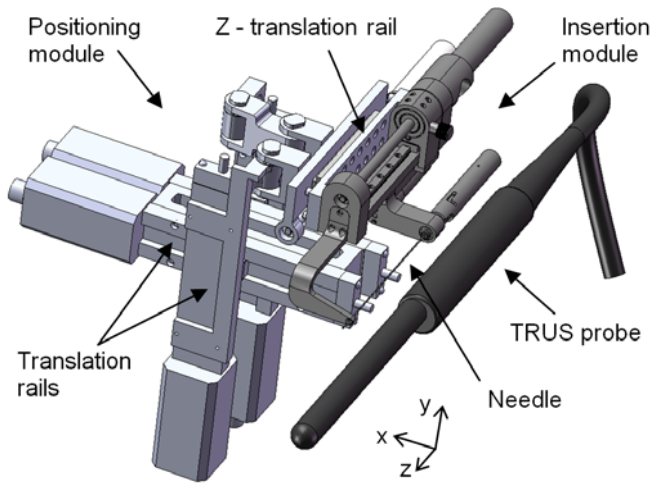

Fig. 2. CAD model of the proposed robotic brachytherapy needle insertion system. The positioning module is shown in light grey and the insertion module in darker grey. The TRUS probe is shown to illustrate the position of the system relative to the operating site. The robot mounting brackets are not shown.

The positioning module consists of two pairs of linear translation rails mounted in the form of a parallelogram-like manipulator and allowing for translation and inclination of the insertion module. For this first laboratory prototype, we chose to use off-the-shelf Zaber T-LLS dovetail slides (Zaber Technologies, Inc.) for the rails, which incorporate rail, carriage, motor and controller in an easy to use and precise package.

Translation of the insertion module in the z-axis allows the needle to be prepositioned near the perineal surface and is achieved by a rail and ball screw combination driven by a brushless DC servomotor (Faulhaber 2057) and a 3.71:1 planetary gear reduction (Z-translation rail in Fig. 2).

The needle insertion module consists of a similar rail, ball screw and servomotor combination that is used to drive the needle during insertion. The needle can be rotated either continuously or by specific amounts in the case of needle-steering. Two novel features are incorporated into this module: the first feature, shown in Fig. 3, is a mechanical release system that disengages the driven ball screw from the needle carriage in case the needle comes in contact with a bone surface, preventing the patient from being harmed. The system functions with an adjustable ball plunger stop whose stiffness is set to release when the axial needle force is greater than the maximum expected tissue-puncture force. It also allows for manual retraction of the needle in case of an electronics malfunction.

The second feature is a needle clamping device that clamps both the needle cannula and stylet for insertion and rotation. Designed specifically for Mick Ripple-Hub needles (Mick Radio-Nuclear Instruments, Inc.), the needle hub and sleeve are manually releasable by pulling and swinging the auto-locking needle holder, as shown in Fig. 4. This allows the clinician to rapidly plug a Mick Applicator or other type of seed dispenser onto the needle. The needle itself is fastened by a removable, sterilized plastic bushing that provides the interface between the sterile needle and the non-sterile elements of the robot. This bushing is linked through an O-ring belt drive to a third brushless DC servomotor (Faulhaber 1536) which drives the rotation of the needle.

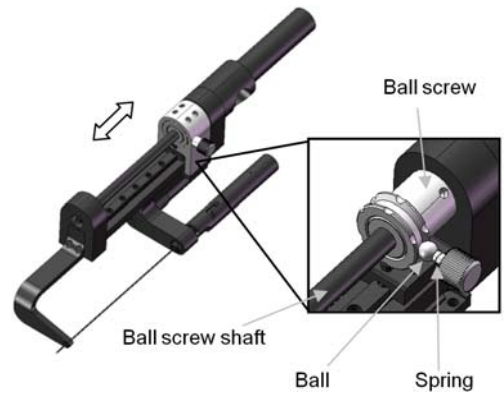

Fig. 3. Mechanical safety release system that disengages the motorized ball screw from the needle carriage in case of bone contact. The system is based on a ball plunger stop consisting of a ball, spring and adjustment screw.

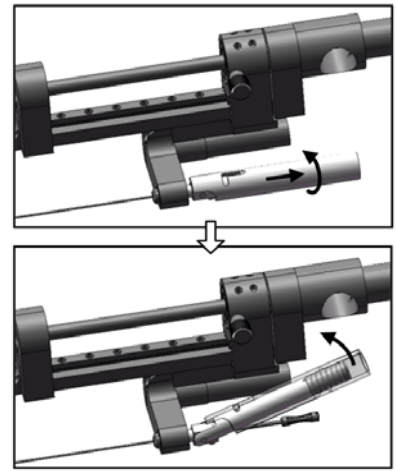

Fig. 4. Illustration showing how the needle hub and sleeve can be rapidly released in preparation for insertion of seeds.

The needle guide at the front of the needle insertion module is sterilizable and exchangeable to accommodate different diameter needles (ex. 18G or 17G). The rest of the needle guide is cleaned but not sterilized. Instead, it is covered by a sterile plastic cap that prevents any non-sterile parts from accidentally touching the sterile zone. The positioning module is covered by sterile drapes, as with the stepper.

V. PRELIMINARY VALIDATION

At the writing of this article, a first prototype was in the process of being built. Preliminary validation of certain elements of the prototype has been made with respect to the design requirements specified.

An experiment was conducted to verify the effect of needle rotation and insertion speed on tissue motion and needle deflection. Using a needle insertion test bench, tests were conducted on various phantoms of homogeneous material simulating soft tissue. Results showed a 25% decrease in the force required to insert a brachytherapy needle when a rotation of 10 rps was applied, as opposed to a static insertion. Additionally, a 15% force decrease was found between a needle inserted at 5 mm/s and one at 1 mm/s. These results will be discussed in further detail in a subsequent article, but they show the potential importance of needle rotation and translation speed on needle-tissue interaction.

A purely mechanical analysis of the prototype has led to the following results: mechanical precision = 0.5 - 1.0 mm; weight = 3.9 kg; workspace = 105 x 105 mm, with inclinations of up to 30°. A more detailed study of the robot's precision, including the effects of needle flexion and tissue motion, is still required.

Validation will be furthered in the coming months upon completion of the prototype. Testing will be done initially on phantom prostates, after which the possibility of cadaver tests will be evaluated.

VI. CONCLUSION

The robotic needle-insertion concept presented above fulfills all the design requirements that we found necessary for providing a beneficial alternative to the manual brachytherapy technique. The design is able to access the same range of horizontal needle positions as the manual template, with the benefit of being able to incline the needles to reach behind the pubic arch or to adjust its reference frame in case of prostate motion. The ability to rotate the needle during insertion will reduce needle-tissue interaction forces and hence increase seed placement precision. In addition, the design accommodates the Mick Applicator seed dispenser while still being open to other dispenser types or even a future automated dispenser.

ACKNOWLEDGMENT

We gratefully acknowledge our clinical team at the CHUG: Pr. Bolla, Pr. Descotes, Dr. Long and J.Y. Giraud, for their invaluable advice and comments.